\pdfoutput=1

\documentclass[11pt]{article}

\usepackage[preprint]{acl}

\usepackage{times}
\usepackage{latexsym}

\usepackage[T1]{fontenc}

\usepackage[utf8]{inputenc}

\usepackage{microtype}

\usepackage{inconsolata}

\usepackage{amsthm}
\usepackage{amsmath}
\usepackage{amssymb}
\usepackage{caption}
\usepackage{subcaption}
\usepackage{enumerate}
\usepackage{multirow}
\usepackage{graphicx} 
\usepackage{url}
\usepackage{booktabs}
\usepackage{tabularx}
\usepackage{colortbl}
\usepackage{algorithm}
\usepackage{algorithmic}
\usepackage{makecell}
\usepackage{tcolorbox}
\usepackage{enumitem}
\usepackage{xcolor}

\definecolor{color1}{HTML}{D0E2BF} 
\definecolor{color2}{HTML}{F7E6D8}

\title{Stepwise Informativeness Search \\
for Effective and Efficient LLM Reasoning}

\author{Siyuan Wang\textsuperscript{\rm 1}\footnotemark[1], Enda Zhao\textsuperscript{\rm 2}\thanks{~Equal contribution.}, 
\textbf{Zhongyu Wei}\textsuperscript{\rm 3}, \textbf{Xiang Ren}\textsuperscript{\rm 1} \\
\textsuperscript{\rm 1}University of Southern California,
\textsuperscript{\rm 2}Tsinghua University,
\textsuperscript{\rm 3}Fudan University. \\
\texttt{sw\_641@usc.edu; zed21@mails.tsinghua.edu.cn} \\
}

\begin{document}
\maketitle
\begin{abstract}
Advances in Large Language Models (LLMs) have significantly improved multi-step reasoning through generating free-text rationales. However, recent studies show that LLMs tend to lose focus over the middle of long contexts. This raises concerns that as reasoning progresses, LLMs may overlook information in earlier steps when decoding subsequent steps, leading to generate unreliable and redundant rationales. To address this, we propose guiding LLMs to generate more accurate and concise step-by-step rationales by (1) proactively referencing information from underutilized prior steps, and (2) minimizing redundant information between new and existing steps. We introduce \textit{stepwise informativeness search}, an inference-time tree search framework incorporating two selection heuristics: grounding-guided selection which prioritizes steps paying higher attention over underutilized steps; and novelty-guided selection which encourages steps with novel conclusions. During rationale generation, we use a self-grounding strategy that prompts LLMs to explicitly reference relevant prior steps to provide premises before deduction at each step. Experimental results on four reasoning datasets demonstrate that our approach improves reasoning accuracy by generating higher-quality rationales with reduced errors and redundancy~\footnote{Code and data are available at \url{https://github.com/SiyuanWangw/Informativeness-Search}.}.
\end{abstract}

\section{Introduction}
Large Language Models (LLMs)~\cite{openai2023gpt4, team2023gemini} have shown remarkable performance in reasoning tasks through Chain-of-Thought (CoT)~\cite{wei2022chain} prompting, which elicits step-by-step rationales to derive answers. However, complex multi-step reasoning remains challenging, particularly for smaller-scale models~\cite{dziri2024faith}. 
\begin{figure}[th!]
    \centering
    \includegraphics[width=1.0\linewidth]{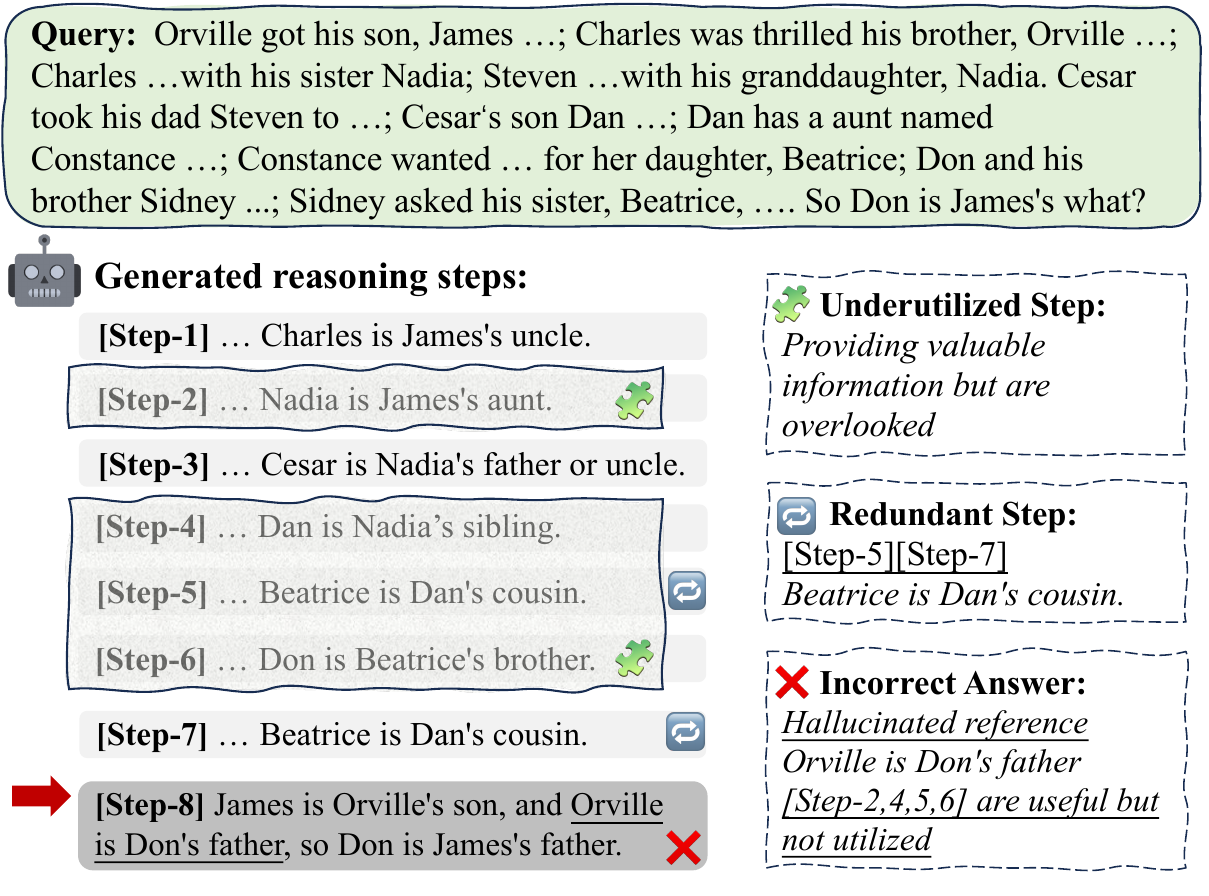}
    \caption{An example illustrating LLMs' difficulty in referencing early-step information (e.g., underutilization of [Step-2,4,5,6]), and the inclusion of redundant steps (e.g., repeated conclusions in [Step-5, 7]). The rightward red arrow indicates the focus is on generating [Step-8] with [Step 1-7] have been generated.
    }
    \label{fig:illustration}
\end{figure}
Recent advances in tree-search algorithms~\cite{wang2024math, yao2024tree, zhang2024accessing} improve this by generating step-level candidates~\footnote{A reasoning step in this paper refers to a sentence in generated rationales, delimited by the end-of-line token ``/n''.} and using scoring mechanisms to select the most promising ones iteratively, thereby improving overall generated rationales. However, they typically rely on domain-specific reward models or more powerful LLMs to assess candidate validity~\cite{luo2024improve}.

Moreover, LLMs tend to focus on leading and recent contexts while losing attention in the middle~\cite{hsieh2024found}. As reasoning progresses, this causes difficulty in referencing useful intermediate conclusions from earlier steps when decoding subsequent ones, leading to unreliable and redundant rationales.
For example, in Fig.~\ref{fig:illustration}, [Step 2,4,5,6] provide useful information for deriving the final answer but are not effectively utilized. This results in redundant steps (e.g., [Step-7] and [Step-5] have repeated conclusions) and incorrect answer (e.g., [Step-8]). 
Consequently, LLMs risk getting trapped in repetitive reasoning loops~\cite{chen2024not} and generating unnecessarily lengthy rationales, increasing the likelihood of cumulative errors~\cite{furuta2024exposing}. 

To address this, we propose to guide LLMs in generating more accurate and concise step-by-step rationales by (1) proactively referencing intermediate conclusions generated from underutilized steps, and (2) minimizing redundancy between new and existing steps. With higher-quality rationales generated, we can improve answer accuracy and reduce decoding costs.
Underutilized steps are those whose intermediate conclusions have been less frequently referenced before the current step, suggesting untapped potential to offer useful information for subsequence reasoning. 
Meanwhile, reducing redundancy across steps can contribute novel information, enabling more efficient exploration of the reasoning space toward final answers.


We introduce \textit{stepwise informativeness search}, an inference-time tree search framework that prioritizes steps based on informativeness, either from leveraging underutilized steps or generating novel content.
The framework follows a stepwise beam search paradigm~\cite{xie2024self}, generating a set of candidate steps in parallel at each iteration. Based on standard cumulative step-level likelihood, it incorporates two heuristics to guide candidate selection. (1) \textit{Grounding-guided selection} identifies underutilized steps by computing each step's reference degree so far to estimate its information gain for subsequent reasoning. Since LLMs naturally assign higher attention to their grounding context~\cite{zhang2023tell}, we prioritize candidate steps with the highest attention scores over underutilized steps.
(2) \textit{Novelty-guided selection} ranks candidates based on the novelty of their intermediate conclusions relative to prior steps. A trigram-based similarity measure filters out highly similar candidates.

To prevent grounding-guided selection from focusing on irrelevant prior steps that may emerge during reasoning, we further introduce a \textit{self-grounding strategy}. This approach elicits LLMs' inherent ability to identify relevant prior steps to provide premises before deduction at each step. This process also extend the possibility of connecting with distant underutilized steps by first specifying their step numbers, and reinforcing the generation of well-supported new steps through explicit grounding.
We implement our informativeness search framework both with and without self-grounding strategy. 
Experimental results across four multi-step reasoning datasets demonstrate the effectiveness of both the informativeness search framework and self-grounding strategy when applied to LLMs of varying families and scales.

Overall, our framework can generate more effective solutions with improved accuracy and fewer tokens. Moreover, the two selection heuristics leverage the model's own outputs and attention scores to intrinsically guide step search, making the approach domain-agnostic and minimizing the need for exhaustive interactions with external scorers or self-evaluation at each decoding step. 

\section{Stepwise Beam Search for Reasoning}
In this work, we formulate multi-step reasoning as a stepwise beam search process considering its generation parallelizability can accelerates search process~\cite{xie2024self}. This contrasts with another common tree-search practice, Monte Carlo Tree Search (MCTS) methods~\cite{feng2023alphazero,zhang2024rest}, which involve extensive rollout simulations and are computationally expensive.

Specifically, at each iteration, the model generates a set of reasoning steps in parallel, each delimited by a special end-of-line token ``/\/n''. A beam of the top $N$ steps are selected according to various criteria, where $N$ is the beam size. Unlike step-level evaluation, stepwise beam search ranks candidates by their cumulative rewards (e.g., likelihood) across the sequence generated so far.

Formally, the generation of a reasoning sequence $R=[s_1, s_2, \dots, s_T]$ with $T$ steps is formulated as 
$$P(R=s_{1:T}|x)=\prod_{t} P(s_t|s_{1:t-1},x),$$
where $s_t$ is the $t$-th step and $x$ is the input query. Stepwise generation and selection are performed with beam size $N$ and sample size $k$ as follows: starting with $N$ sequences at step $t-1$, it generates $k$ continuations from $P(s_t|s_{1:t-1},x)$ for each sequence $s_{1:t-1}$, forming a candidate set $C_t$ containing $Nk$ reasoning chains of length $t$. The top $N$ sequences are then selected based on a scoring criteria $\phi(C_t, \mathrm{\gamma}(\cdot))=\{s^1, s^2, \dots, s^N\}$. $\phi$ is the selection function (e.g., $topk(\cdot)$) and $\mathrm{\gamma}(s_{1:t})$ evaluates the sequence so far $s_{1:t}$. Initially, given only an input $x$, we generate $Nk$ candidates.



\begin{figure*}[th!]
    \centering
    \includegraphics[width=1.0\linewidth]{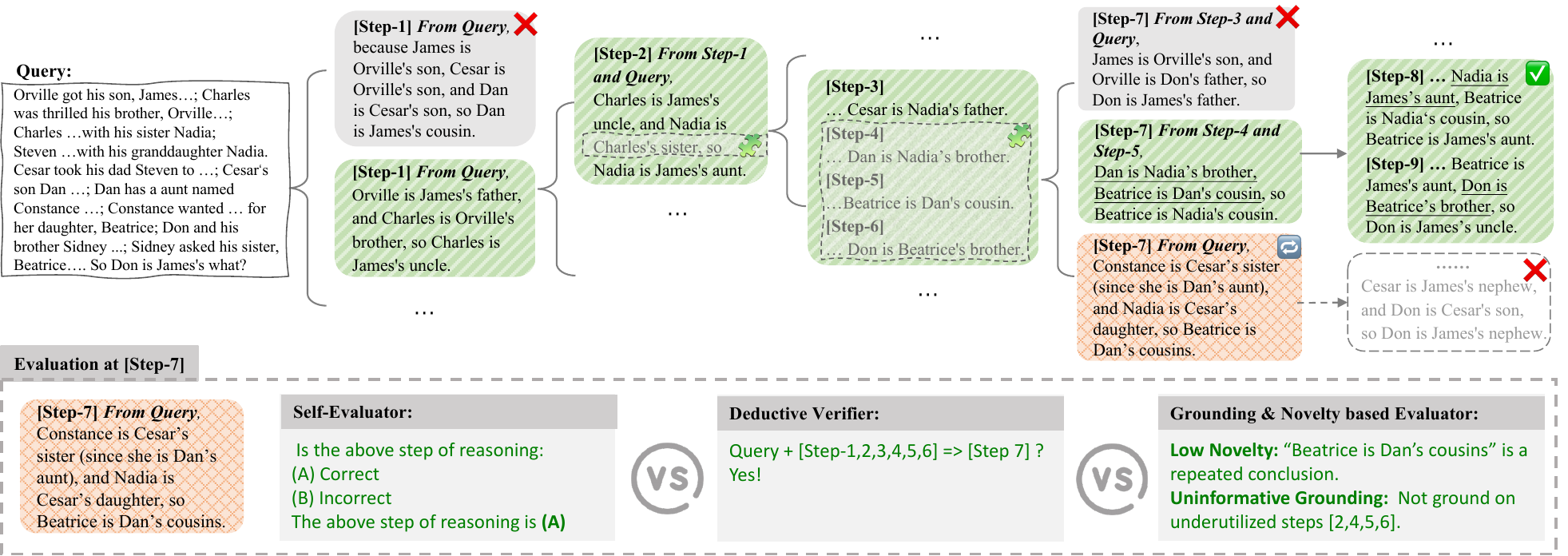}
    \caption{\textbf{Upper:} Overview of our informativeness search framework, illustrated with beam size of 1. Green diagonal-striped blocks represent selected steps while gray blocks are discarded. Cross marks indicate incorrect deductions, and the orange crosshatched block highlights a redundant step that may lead to errors. Italics illustrate our self-grounding strategy. \textbf{Bottom:} While previous methods would accept this redundant [Step-7] as logically valid, our framework filters it out based on its low novelty and poor grounding on underutilized steps.}
    \label{fig:workflow}
\end{figure*}

A standard scoring criteria is the cumulative likelihood of a sequence, defined as: $\mathrm{\gamma}_L(s_{1:t})=\log\prod_{t} P(s_t|s_{1:t-1}, x)$. Alternative scoring functions $\mathrm{\gamma}(s_{1:t})$ are employed in self-evaluation~\cite{xie2024self} and deductive beam search~\cite{zhu2024deductive}. 
The former prompts the backend LLM to provide a correctness score $\mathrm{\gamma}_c(s_t)$ to assess whether $s_{t}$ is correct given $s_{1:t-1}$, which is then combined with likelihood: $\mathrm{\gamma}_{E}(s_{1:t})=\log \prod_t P(s_t|s_{1:t-1}, x) \ \mathrm{\gamma}_c(s_t).$
The latter trains an external deductive verifier $f$ to assess whether each step $s_{t}$ is logically entailed by previous contexts, and replaces the sequence likelihood with a cumulative deductive score: 
$\mathrm{\gamma}_{D}(s_{1:t})=\prod_t f(\text{entails}|s_t, s_{1:t-1}, x).$

While these methods improve performance, they require additional annotations or prompts to obtain domain-specific scoring models. They also incur interaction overhead by waiting for scorer response at each decoding step, yet failing to address aforementioned grounding and redundancy challenges.

\section{Informativeness Search Framework}
Unlike iteration-based scoring functions described above, we introduce stepwise informativeness search framework with two scoring heuristics that utilize model's intrinsic outputs and attention scores. This reduces reliance on off-the-shelf scorers and iterative interactions during decoding. It prioritizes steps based on informativeness, assessed by grounding-guided and novelty-guided heuristics that determine whether new decoded steps ground on underutilized steps and generate novel content.

\subsection{Grounding-Guided Selection}
To ground each deduction upon underutilized steps to maximally leverage useful intermediate information, we design an algorithm to identify underutilized ones among all prior steps. The candidate sequences, denoted as $C_t=\{s_{1:t}^1, s_{1:t}^2, \dots, s_{1:t}^{Nk}\}$, are then evaluated and selected based on whether each current step $s^i_{t}$ is well derived from its corresponding underutilized steps.

\paragraph{Identifying Underutilized Steps} At each reasoning step, underutilized steps are those referenced less frequently up to that point, offering higher untapped potential for contributing information to subsequent reasoning. At the current step $s_t$, the immediately preceding step $s_{t-1}$ is by default considered underutilized since it represents the most recent addition to the reasoning path. For additional underutilized steps, we perform a backward traversal from step $s_{t-2}$ to $s_1$, calculating the reference degree of each step to assess its information gain to subsequent reasoning. 

Specifically, for each prior step $s_{j} \in \{s_{t-2}, ..., s_2, s_1\}$, we first extract its intermediate conclusion $c_j$ by segmenting it using special clause delimiters (e.g., ``so'', ``thus'' and commas). We then compare $c_j$ with each subsequent step $s_m\in \{s_{j+1}, \dots, s_{t-1}\}$ before the current step using a trigram-based similarity measure. The information gain of $s_{j}$ is computed as follows: 
\begin{align*}
    \mathrm{InfoGain}(s_{j}) = 1 - \max\limits_{m \in {j+1, \dots, t-1}}\mathrm{Sim}_{\mathrm{tri}}(c_{j}, s_{m})
\end{align*}
We classify a prior step as underutilized if its information gain exceeds a predefined threshold $\tau$. The set of underutilized steps at step $t$ is:
\begin{align*}
    \mathcal{I}_t = \{s_{t-1}\} \ \cup \ \{s_j \mid \mathrm{InfoGa}&\mathrm{in}(s_j) > \tau\}, \\
    &j \in \{1, \dots, t-2\} 
\end{align*}

\paragraph{Grounding on Underutilized Steps} 
After identifying the set of underutilized steps $\mathcal{I}^i_t$ for each candidate sequence $s^i_{1:t}$ in the candidate set $C_t=\{s^1, s^2, \dots, s^{Nk}\}$ (with subscripts omitted for simplicity), we prioritize candidates that more effectively ground their reasoning in $s^i_t$ upon their respective underutilized steps. 

LLMs typically assign higher attention scores to their grounding context~\cite{zhang2023tell}. We leverage attention scores to evaluate how well each candidate focuses on and utilizes its identified underutilized steps $\mathcal{I}^i_t$ when constructing step $s^i_t$ in the reasoning path. We specifically compute the attention score of $s^i_t$ over $\mathcal{I}^i_t$ as $\mathrm{\gamma}_a(s^i_t)$ by applying mean pooling across all tokens in $s^i_t$ and the highly attended tokens within $\mathcal{I}^i_t$. 
We then integrate this attention-based measure into the original cumulative likelihood scoring function to obtain an grounding-enhanced score:
\begin{align*}
    \mathrm{\gamma}_G(s_{1:t}) & = \mathrm{\gamma}_{L}(s_{1:t}) + \alpha \cdot \mathrm{\gamma}_a(s_t)
\end{align*}
where $\mathrm{\gamma}_L(s_{1:t}) = \log \prod_t P(s_t|s_{1:t-1}, x)$ and $\alpha$ is a weighted hyperparameter. Then $N$ candidates are selected from $C_t=\{s^1, s^2, \dots, s^{Nk}\}$ with the highest $\mathrm{\gamma}_G(s_{1:t})$. We validate this attention-based operation in Sec.~\ref{sec:attention_validation} by analyzing the consistency between highly attended content and actual grounded information.

\subsection{Novelty-Guided Selection}
\label{sec:novelty-guidance}
To reduce redundancy across multiple intermediate steps, we assess the conclusion novelty of each newly generated step $s^i_t$ in a candidate sequence $s^i_{1:t}$, and select candidates with higher novelty.
We extract intermediate conclusions from $s^i_t$ and all its prior steps $\{s^i_1, \dots, s^i_{t-1}\}$ by segmenting the corresponding sentences using special clause delimiters (e.g., ``so'', ``thus'' and commas), forming a set of conclusions $\{c^i_1, \dots, c^i_{t-1}, c^i_{t}\}$. We then calculate the trigram-based similarity between the newly generated conclusion $c^i_{t}$ and all preceding conclusions $\{c^i_1, \dots, c^i_{t-1}\}$. The novelty score of $s^i_t$ is then obtained as follows:
\begin{align*}
    N(s^i_t) = 1 - \max\limits_{j \in {1, \dots, t-1}}\mathrm{Sim}_{\mathrm{tri}}(c^i_{t}, c^i_{j})
\end{align*}
where $\mathrm{Sim}_{\mathrm{tri}}(\cdot, \cdot)$ measures trigram-based similarity. To incorporate novelty into candidate selection, we calibrate the grounding-enhanced scoring function with novelty score. At step $t$, candidates with low-novelty conclusions (i.e., $N(s_t) \leq \theta$) are filtered out, retaining only diverse and meaningful candidates. The adjusted scoring function is defined as below, where $\theta$ is a predefined threshold.
\begin{align*}
    \mathrm{\gamma}_{N}(s_{1:t}) = \begin{cases} 
    \mathrm{\gamma}_G(s_{1:t}), & \text{if} N(s_t)>\theta, \\ 
    -100, &\text{otherwise}.
    \end{cases}
\end{align*}

\subsection{Self-Grounding Strategy}
To handle irrelevant steps that may arise during reasoning generation and prevent grounding-guided selection from focusing on irrelevant prior steps, especially when contexts contain distracting information, we introduce a self-grounding strategy. This approach leverages LLMs' inherent ability to anchor reasoning in relevant prior information, either from prior steps or the input query, that serve as necessary premises for each new deduction. The strategy explicitly prompts LLMs to reason step by step, structuring each step in the format:
$$
\texttt{``[Step-i] From <source>, <deduction>.''}
$$ 
where ``<source>'' refers to either relevant prior steps or the input query that provide premises for deducing new conclusions in ``<deduction>''. For example, ``\textit{[Step-1] From Query, we know ...}'',
``\textit{[Step-2] From Step-1 and Query, we know ...}'' and ``\textit{[Step-3] From Step-1 and Step-2, because ...}''.
This explicit step-grounding process ensures that each new step directly builds upon established information, maintaining logical coherence while minimizing irrelevant or unsupported conclusions. Moreover, explicitly referencing step numbers facilitates connections with distant underutilized steps. Further details on the prompts and few-shot demonstrations are provided in Appendix~\ref{prompts}.

\begin{table*}[!th]
    \centering
    \setlength\tabcolsep{5pt}
    \resizebox{1.0\textwidth}{!}{
    \begin{tabular}{cc|cccc|c}
    \toprule
    Models & Methods & FOLIO & ProofWriter & MMLU-Pro & GPQA-Diamond & Avg.\\
    \bottomrule
    \addlinespace[1pt]
    \multirow{10}{*}{Llama3.2-3B-Instruct} & 
    \textit{Few-shot CoT} & 38.73\% & 40.00\% & 28.57\% & 21.72\% & 32.25\% \\
    & \textit{Self-Grounding CoT} & 45.59\% & 43.33\% & 28.57\% & 22.73\% & 35.06\% \\
    & \textit{Best-of-N} & 45.59\% & 37.00\% & 30.00\% & 22.73\%  & 33.83\%\\ 
    & \textit{Self-Consistency} & 46.57\% & 47.67\% & 29.64\% & 22.73\% & 36.65\% \\
    \addlinespace[1pt]
    \cline{2-7}
    \addlinespace[1pt]
    & \textit{Tree-of-Thought} & 44.12\% & 44.17\% & 26.43\% & 22.73\% & 34.36\%  \\
    & \textit{Self-Eval Beam Search} & 45.10\% & 47.00\% & 30.71\%& 19.19\%& 35.50\% \\
    & \textit{Deductive Beam Search} & 48.04\% & 38.17\% & 25.71\% & 24.75\% & 34.17\% \\ 
    & \textit{MCTS + Math-PRM} & / & / & 26.07\% & 22.22\% & / \\ 
    & \cellcolor[HTML]{E1E1E1} \textit{Informativeness Search} & \cellcolor[HTML]{E1E1E1} 46.57\% & \cellcolor[HTML]{E1E1E1} 50.33\% & \cellcolor[HTML]{E1E1E1} 33.57\% & \cellcolor[HTML]{E1E1E1} \bf 27.27\% & \cellcolor[HTML]{E1E1E1} 39.44\% \\
    & \cellcolor[HTML]{E1E1E1} \textit{Informativeness Search (w/ SG)} & \cellcolor[HTML]{E1E1E1} \bf 51.96\% & \cellcolor[HTML]{E1E1E1} \bf 53.67\% & \cellcolor[HTML]{E1E1E1} \bf 33.93\% & \cellcolor[HTML]{E1E1E1} 24.24\bf \% & \cellcolor[HTML]{E1E1E1} \bf 40.95\% \\
    \midrule
    \multirow{10}{*}{Llama3-8B-Instruct} & 
    \textit{Few-shot CoT} & 54.90\% & 55.33\% & 37.50\% & 29.29\% & 44.25\% \\
    & \textit{Self-Grounding CoT} & 55.39\% & 57.00\% & 38.57\% & 30.30\% & 45.32\% \\
    & \textit{Best-of-N} & 56.86\% & 50.00\% & 39.29\% & 30.30\% & 44.11\%\\ 
    & \textit{Self-Consistency} & 57.84\% & 60.17\% & 39.29\% & 31.31\% & 47.15\% \\
    \addlinespace[1pt]
    \cline{2-7}
    \addlinespace[1pt]
    & \textit{Tree-of-Thought} & 55.88\% & 53.33\% & 39.29\% & 27.78\% & 44.07\%  \\
    & \textit{Self-Eval Beam Search} & 59.31\% & 56.17\% & 35.00\% & 29.80\% & 45.07\% \\
    & \textit{Deductive Beam Search} & 54.90\% & 48.83\% & 37.50\% & 27.78\% & 42.25\% \\ 
    & \textit{MCTS + Math-PRM} & / & / & 27.14\% & 28.28\% & / \\ 
    & \cellcolor[HTML]{E1E1E1} \textit{Informativeness Search} & \cellcolor[HTML]{E1E1E1} 58.33\% & \cellcolor[HTML]{E1E1E1} 61.33\% & \cellcolor[HTML]{E1E1E1} 40.00\% & \cellcolor[HTML]{E1E1E1} 33.33\% & \cellcolor[HTML]{E1E1E1} 48.25\% \\
    & \cellcolor[HTML]{E1E1E1} \textit{Informativeness Search (w/ SG)} & \cellcolor[HTML]{E1E1E1} \bf 59.80\% & \cellcolor[HTML]{E1E1E1} \bf 62.00\% & \cellcolor[HTML]{E1E1E1} \bf 40.71\% & \cellcolor[HTML]{E1E1E1} \bf 35.35\% & \cellcolor[HTML]{E1E1E1} \bf 49.46\% \\
    \bottomrule
    \end{tabular}
    }
    \caption{Experimental results (accuracy \%) of different methods on Llama3.2-3B-Instruct and LLaMA3-8B-Instruct. \textit{SG} denotes the \textit{Self-Grounding} strategy. Shaded rows present results from our proposed method.}
    \label{overall_result}
\end{table*}
\section{Experiments}
\subsection{Setup}
\textbf{Datasets} We evaluate our framework on four multi-step reasoning datasets: FOLIO~\cite{han2022folio}, ProofWriter~\cite{tafjord2020proofwriter}, MMLU-Pro~\cite{wang2024mmlu} and GPQA~\cite{rein2023gpqa}. FOLIO and ProofWriter focus on deductive reasoning, requiring 1-8 and 1-6 reasoning steps respectively, with test sets of 204 and 600 cases. MMLU-Pro covers 14 domains, including math, physics, chemistry, engineering, law, and psychology, from which we uniformly sample 280 cases. GPQA specializes in biology, physics, and chemistry, and we use its Diamond subset containing 198 expert-answered but non-expert-failed questions.

\vspace{1mm}
\noindent\textbf{Baselines} We evaluate against both sequence-level CoT methods and step-level search methods. Sequence-level methods include: (1) Few-shot CoT~\cite{wei2022chain} performs step-by-step reasoning.
(2) Self-Grounding CoT is our proposed self-grounding strategy without search. 
(3) Best-of-N~\cite{lightman2023let} samples $Nk$ rationales and selects the best via LLM self-evaluation as we lack general reward models for diverse tasks. (4) Self-Consistency~\cite{wang2022self} samples $Nk$ rationales and uses majority voting for the final answer.
Step-level methods include: (5) Tree-of-thought~\cite{yao2024tree} performs breadth-first tree search with self-evaluation at each step.
(6) Self-Eval Beam Search~\cite{xie2024self} and (7) Deductive Beam Search~\cite{zhu2024deductive} both use stepwise beam search, with the former relying on self-evaluation and the latter on deductive scoring trained on synthesized datasets. (8) MCTS~\cite{zhang2024rest} where we use the minimum score across all steps from Qwen2.5-Math-PRM-7B~\cite{zhang2025lessons} to evaluate simulated solutions. As this is a mathematical PRM, we report MCTS results only on MMLU-Pro and GPQA-Diamond. We evaluate our informativeness search with and without the self-grounding (SG) strategy.

\vspace{1mm}
\noindent\textbf{Implementation Details} We evaluate our method and baselines on Llama3.2-3B-Instruct and Llama3-8B-Instruct, using a two-shot prompting strategy with a 1024-token generation limit. We set $N=3$ and $k=2$ for all stepwise beam search methods. The weighted parameter $\alpha$ is set to 2 and the threshold $\tau$ to 0.7. $\theta$ is set to 0.5 for FOLIO and ProofWriter, 0.4 for MMLU-Pro and GPQA-Diamond. Further details and search configurations are provided in Appendix~\ref{sec:implementation}. 

\begin{figure*}[th!]
    \centering
    \includegraphics[width=0.94\linewidth]{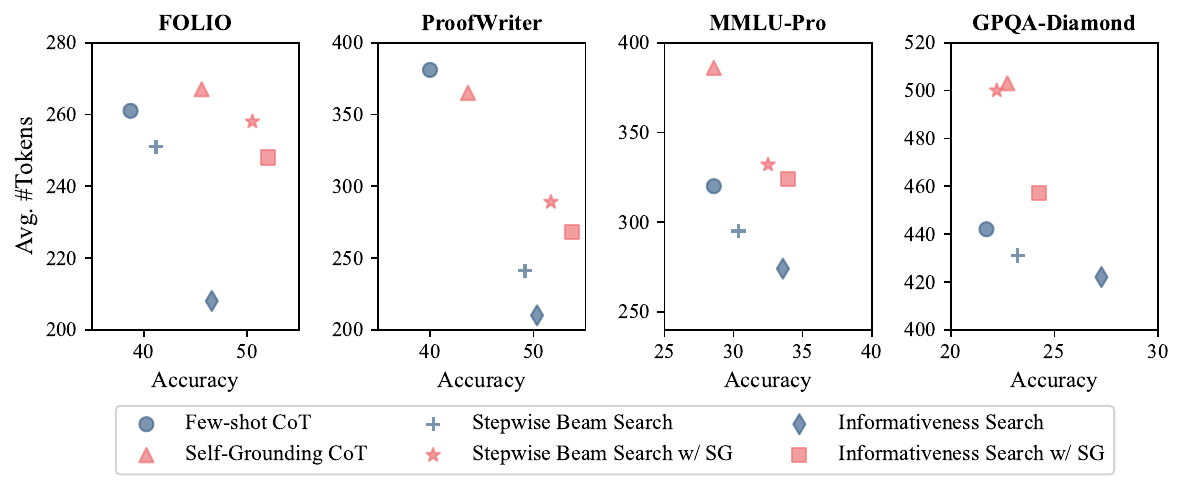}
    \caption{Accuracy and average token count (Avg. \# Tokens) of final predicted rationales using different methods on Llama3.2-3B-Instruct.}
    \label{fig:avg_tokens}
\end{figure*}

\subsection{Main Results}
Table~\ref{overall_result} presents the overall performance comparison across four benchmark datasets. Our method consistently outperforms all baseline methods across both deductive and diverse reasoning datasets when implemented with either Llama3.2-3B-Instruct or Llama3-8B-Instruct. This demonstrates the general superiority of our informativeness search framework and self-grounding strategy. Notably, our method yields more substantial improvements on Llama3.2-3B-Instruct, suggesting its particular effectiveness in enhancing reasoning for lower-performing models. Additionally, self-grounding further enhances informativeness search, except when using Llama3.2-3B-Instruct on GPQA-Diamond. We attribute this to Llama3.2-3B-Instruct's inability to perform self-grounding effectively for the challenging GPQA-Diamond task. Step-level methods like tree-of-thought, deductive beam search and MCTS show moderate performance due to their reliance on specialized reward model or verifiers, limiting their generalizability. In contrast, informativeness search is broadly applicable without requiring task-specific customization. 


\subsection{Efficiency Analysis}
\paragraph{Average Rationale Length} We analyze the average token count of final predicted rationales using different methods on Llama3.2-3B-Instruct to examine the relationship between rationale length and accuracy. As shown in Table~\ref{fig:avg_tokens}, our method generates shorter rationales with fewer tokens than few-shot CoT and stepwise beam search while achieving higher accuracy, both with and without the self-grounding strategy. Notably, our approach exhibits greater token reduction in deductive reasoning, correlating with more significant performance improvements. We attribute this to our informativeness search framework can effectively reduce redundancy by combining grounding-guided and novelty-guided selection. This minimizes cumulative errors and prevents circular reasoning loops, ultimately leading to better performance.

\begin{figure}
    \centering
    \includegraphics[width=0.93\linewidth]{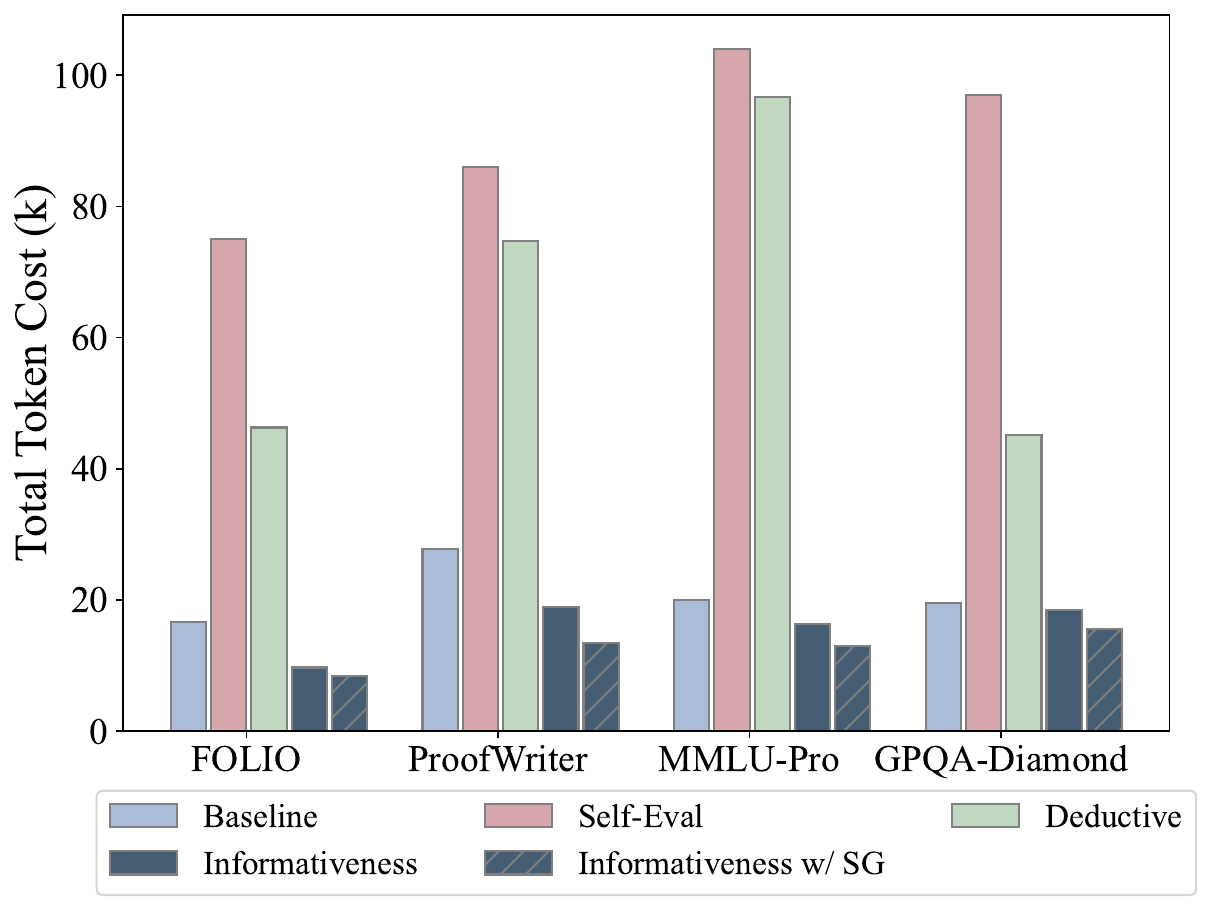}
    \caption{Total token costs ($\times k$ tokens) of different stepwise beam search methods. Baseline refers to stepwise beam search using only cumulative likelihood scoring.}
    \label{fig:total_cost}
\end{figure}

\paragraph{Total Token Cost} We further analyze the total token consumption following~\cite{xie2024self}, including all candidate steps during the stepwise beam search process for all methods involving stepwise beam search. As shown in Table~\ref{fig:total_cost}, our method exhibits superior inference efficiency, reducing token usage compared to the baseline and other beam search methods. 
Specifically, both informativeness search and self-grounding progressively reduce token budget compared to baseline stepwise beam search. The high costs of self-eval and deductive beam search stem from additional interactions for obtaining evaluation feedback after each step. Moreover, deductive beam search requires additional computational resources for training a domain-specific deductive verifier. 

\subsection{Results on Additional LLMs}
To further validate the broad effectiveness of our method, we implement it on Phi-4~\cite{abdin2024phi}, a 14B-parameter model from a different model family, and DeepSeek-R1-Distill-Llama-8B~\cite{guo2025deepseek}, a slow-thinking Llama3-8B variant distilled from DeepSeek-R1. We evaluate performance on FOLIO, ProofWriter, and MMLU-Pro, comparing against few-shot CoT, self-grounding, and self-consistency baselines using corresponding backbones. A one-shot prompting strategy is used with $N=3$ and $k=1$, and we extend the generation limit to 2048 tokens to accommodate long CoT from R1-Distill-Llama-8B.
As shown in Table~\ref{phi4-result}, our framework consistently improves performance on more powerful LLMs, though self-grounding fails on R1-Distill-Llama-8B, as it learns to generate free-form CoT and struggles to follow a structured response format. Despite this, our informativeness search still yields significant improvements, notably reducing redundant tokens in final rationales (Table~\ref{r1_llama_tokens}).
This aligns with DeepSeek-R1's over-thinking problems as pointed by~\cite{chen2024not, cuadron2025dangeroverthinkingexaminingreasoningaction}.
These results, along with Table~\ref{overall_result} demonstrate our method's robustness across models. 
\begin{table}[!th]
    \centering
    \setlength\tabcolsep{2pt}
    \resizebox{0.48\textwidth}{!}{
    \begin{tabular}{c|ccc}
    \toprule
    Method &  FOLIO & ProofWriter & MMLU-Pro \\
    \midrule
    \rowcolor[HTML]{E1E1E1} \multicolumn{4}{c}{\textit{Phi-4}} \\
    Few-shot CoT & 73.67\% & 72.55\% & 71.79\% \\
    Self-Grounding CoT & 73.50\% & 72.06\% & 72.14\% \\
    Self-Consistency & 71.17\% & 72.55\% & 72.50\%  \\
    \midrule
    Informativeness Search w/ SG & 76.67\% & 77.94\% & 72.86\%\\
    \midrule
    \rowcolor[HTML]{E1E1E1} \multicolumn{4}{c}{\textit{DeepSeek-R1-Distill-Llama-8B}} \\
    Few-shot CoT & 61.76\% & 48.67\% & 38.57\% \\
    Self-Grounding CoT & 53.92\% & 38.17\% & 35.36\% \\
    Self-Consistency & 62.25\% & 63.50\% & 46.07\%  \\
    \midrule
    Informativeness Search & 70.10\% & 66.50\% & 47.50\%\\
    \bottomrule
    \end{tabular}
    }
    \caption{Results on Phi-4 and R1-Distill-Llama-8B.}
    \label{phi4-result}
\end{table}
\begin{table}[!th]
    \centering
    \vspace{-3mm}
    \setlength\tabcolsep{3pt}
    \resizebox{0.45\textwidth}{!}{
    \begin{tabular}{c|ccc}
    \toprule
    Method &  FOLIO & ProofWriter & MMLU-Pro \\
    \midrule
    Few-shot CoT & 1105 & 1861 & 1636 \\
    Informativeness Search & 588 & 1023 & 1001 \\
    \bottomrule
    \end{tabular}
    }
    \caption{Average token count of the final predicted reasoning paths from R1-Distill-Llama-8B.}
    \label{r1_llama_tokens}
\end{table}

\section{Further Analysis} 

\subsection{Ablation Study}
To investigate the contribution of different components in our method, we conduct an ablation study using LLama3.2-3B-Instruct on FOLIO and MMLU-Pro datasets. Starting with stepwise beam search as our baseline, we progressively add: (1) novelty-guided selection heuristic, (2) grounding-guided selection heuristic (forming our \textit{Informativeness Search Framework}), and (3) self-grounding strategy (resulting in \textit{Informativeness Search w/ SG}). 
As shown in Table~\ref{ablation_study}, incorporating each selection heuristic and self-grounding strategy incrementally improves performance, finally yielding our best-performing informativeness search framework with self-grounding. Notably, novelty-based selection proves especially effective on FOLIO, suggesting that deductive reasoning is more susceptible to redundant step generation. Furthermore, self-grounding achieves more significant improvement on deductive reasoning where contexts contain verbally similar but irrelevant information.
\begin{table}[!th]
    \centering
    \setlength\tabcolsep{4.5pt}
    \resizebox{0.48\textwidth}{!}{
    \begin{tabular}{lcc}
    \toprule
    Methods & FOLIO & MMLU-Pro \\
    \midrule
    Stepwise Beam Search & 41.18\% & 30.36\% \\
    \ + Novelty-Guided Heuristic & 45.10\% & 32.14\% \\
    \ + Grounding-Guided Heuristic & 46.57\% & 33.57\% \\  
    \ + Self-Grounding Strategy & 51.96\% & 33.93\% \\
    \bottomrule
    \end{tabular}
    }
    \caption{Ablation study using LLaMA3.2-3B-Instruct. 
    }
    \label{ablation_study}
\end{table}

\subsection{Redundant Step Analysis}
\begin{figure}[ht]
    \centering
    \includegraphics[width=0.94\linewidth]{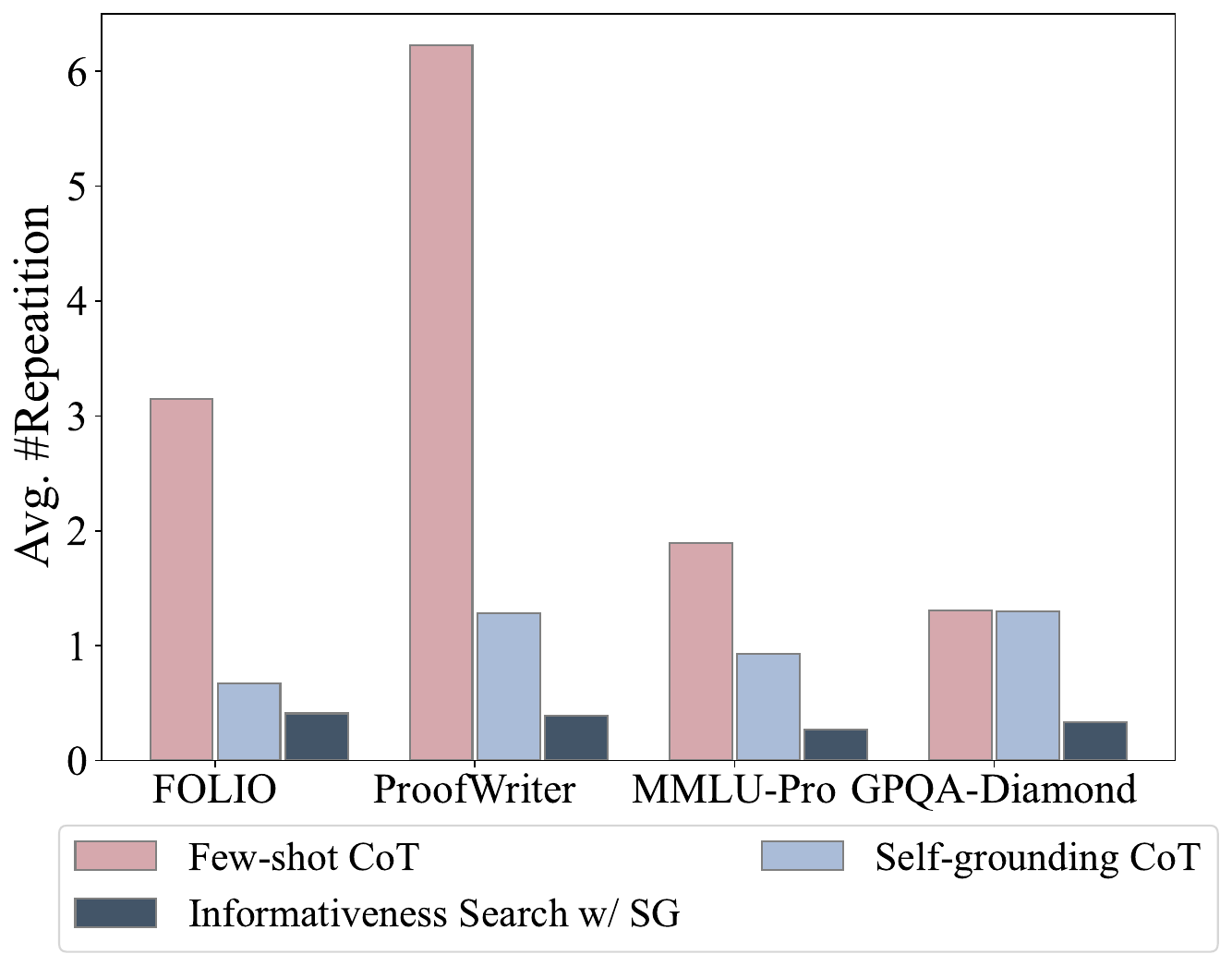}
    \caption{Average count of redundant steps whose conclusions have over 70\% tri-word overlap with any previous conclusions in the same rationale.}
    \vspace{-4mm}
    \label{fig:repetitive_deduction}
\end{figure}

In complex multi-step reasoning tasks, LLMs tend to generate repeated intermediate conclusions, either from same or different premises, which can trap reasoning in circular loops. For detailed investigation, we measure the average number of repeated conclusions across steps per rationale generated by our method compared to few-shot CoT and self-grounding CoT baselines using LLama3.2-3B-Instruct. Specifically, we split rationales into steps using end-of-line token ``/\/n'' and extract intermediate conclusions based on special clause delimiters as operated in Sec.~\ref{sec:novelty-guidance}. A step is considered redundant if its conclusion shares over 70\% tri-word overlap with any previous conclusions in the same rationale. 
As shown in Figure~\ref{fig:repetitive_deduction}, LLMs exhibit a pronounced tendency to produce redundant steps, particularly in
deductive reasoning tasks. This occurs because deductive contexts often contain verbally similar information, causing LLMs to lose
track of logical progression and become stuck in circular reasoning. In contrast, our self-grounding strategy and informativeness search substantially reduce redundant steps, enabling more effective and efficient multi-step reasoning.

\subsection{Validity of Attention-Based Selection}
\label{sec:attention_validation}
\begin{figure}[ht]
    \centering
    \includegraphics[width=0.98\linewidth]{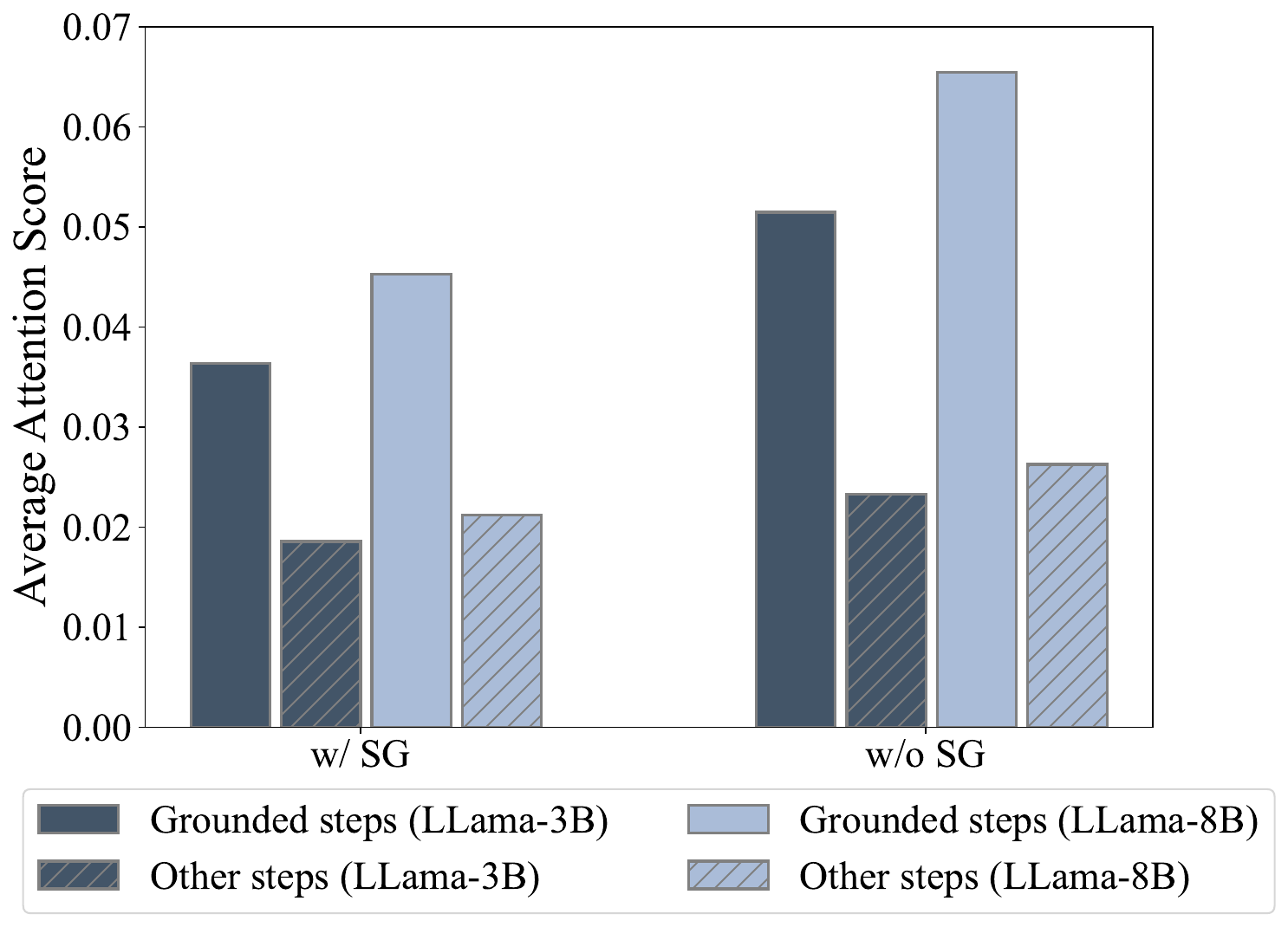}
    \caption{Average attention on grounded and other steps.}
    \label{fig:grounded_scores}
\end{figure}

To validate our attention-based implementation in grounding-guided selection, we examine whether LLMs naturally assign higher attention to grounded steps than other steps. Using the CLUTRR dataset, which provides well-annotated reasoning paths, we conduct a teacher-forcing analysis where all previous ground-truth steps are fed into the model to prompt the next step. We then compute the average attention score over both grounded and non-grounded steps. This analysis is performed both with and without self-grounding, using Llama3.2-3B-Instruct and Llama3-8B-Instruct. As shown in Fig.~\ref{fig:grounded_scores}, LLMs exhibit significantly higher attention over grounded steps. 
This demonstrates the consistency of LLMs' attention patterns and their grounding behavior, and confirms the validity of our attention-based implementation.




\section{Related Work}
LLMs~\cite{openai2023gpt4, touvron2023llama, abdin2024phi, guo2025deepseek} have demonstrated remarkable performance across diverse tasks. Chain-of-Thought (CoT)~\cite{wei2022chain, zhou2022least} prompting has emerged as an effective strategy for generating step-by-step rationale to derive answers. However, for complex multi-step reasoning problems, LLMs often underutilize critic information from earlier steps as rationale getting longer due to their tendency to lose focus on middle-context information~\cite{peysakhovich2023attention, junqing2023never, hsieh2024found}. Additionally, they frequently generate redundant steps with repeated conclusions, leading to repetitive reasoning loops and error accumulation~\cite{dziri2024faith, furuta2024exposing}. These difficulties are especially pronounced in smaller-scale LLMs with limited reasoning capacity~\cite{fu2023specializing}.
An intuitive method is to prompt LLMs for more concise outputs. However, LLMs often struggle to maintain output quality under length constraints, and simple prompting alone fails to resolve grounding and redundancy issue~\cite{nayab2024concise, han2024token}. 

This inspires generating multiple rationales and determine the most likely solution using majority voting~\cite{wang2022self} or best-of-N~\cite{wang2024math}. However, they are computationally expensive due to the exponentially growing search space when integrating diverse solutions.
To reduce the search space, recent studies have applied tree search techniques with scoring mechanisms to prioritize promising candidates at each step, such as stepwise beam search~\cite{xie2024self}, Tree-of-Thought~\cite{yao2024tree}, and Monte Carlo Tree Search~\cite{jiang2024technical, feng2023alphazero, zhang2024rest}. While effective, they face practical limitations, relying on extensive rollouts~\cite{wang2024math, wang2024q} and intensive annotations~\cite{lightman2023let} for training specialized reward models. Additionally, they introduce latency due to interactions with external or self-evaluators during autoregressive decoding~\cite{xie2024self, yao2024tree}, and fail to address the grounding and redundancy issues we focus on in this work.

\section{Conclusion}
In this work, we address the challenge of LLMs losing focus on intermediate steps during multi-step reasoning, which can lead to unreliable and redundant rationales. To mitigate this issue, we propose an inference-time tree search framework incorporating grounding-guided and novelty-guided selection heuristics, that enhances rationale generation by proactively grounding underutilized prior steps and minimizing redundant conclusions between reasoning steps. We additionally employ a self-grounding strategy, prompting LLMs to explicitly reference relevant prior steps before making deductions. Experimental results demonstrate that our method improves reasoning accuracy by generating higher-quality rationales with fewer errors and reduced redundancy. 

\section*{Limitations}
Our work has several limitations to address in future research. 
First, our experiments primarily focus on four multi-step reasoning datasets covering deductive and diverse-discipline reasoning. Expanding to a broader range of tasks and datasets will further validate our framework’s effectiveness.
Second, due to computational constraints, our main experiments operate within a limited search space with beam size 3 and sample size 2, and use LLM backbones of at most 14B parameters. Future work can explore larger search spaces and more powerful LLMs to further unlock the potential of our framework. Finally, while our method currently relies solely on stepwise beam search with standard cumulative likelihood, incorporating our selection heuristics with other scoring mechanism, such as self-evaluation and process reward models, as well as other tree-search algorithms like MCTS could be potential future work.

\bibliography{custom}

\appendix

\newpage
\section{Implementation Details}
\label{sec:implementation}

\subsection{Baseline Details}
For Best-of-N and Self-Consistency, we adopt a sampling configuration with temperature $T = 1.0$ and top-$40$ token truncation. For tree-of-thought (ToT) and self-eval beam search (Self-Eval BS), we prompt LLMs to conduct self-evaluation. For deductive beam search that provide a general verifier checkpoint and two data subsets for training a commonsense and a mathematical verifier, we select the best-performing verifier for each dataset. Specifically, we use the general or commonsense verifier for FOLIO, ProofWriter, and MMLU-Pro, and the general or mathematical verifier for GPQA.
For MCTS which operates in a iterative four-stage manner: selection, expansion, simulation and backprogation, we use the minimum score across all steps from Qwen2.5-Math-PRM-7B~\cite{zhang2025lessons} to evaluate simulated rollout. 

\subsection{Varying Search Configurations}
For step-level candidate generation in stepwise beam search, we explore both temperature sampling and tokenwise beam search. 
As shown in Table~\ref{candidate_generation}, our method with grounding and novelty-guided selection consistently outperforms stepwise beam search baseline (with cumulative likelihood scoring), regardless of whether self-grounding is applied. Additionally, tokenwise beam search for candidate generation yields slightly better performance than temperature sampling. 
\begin{table}[!th]
    \centering
    \setlength\tabcolsep{4pt}
    \resizebox{0.49\textwidth}{!}{
    \begin{tabular}{lcc}
    \toprule
    Methods & FOLIO & MMLU-Pro \\
    \midrule
    \rowcolor[HTML]{E1E1E1} \multicolumn{3}{c}{\textit{Beam Search}} \\
    Stepwise Beam Search & 41.18\% & 30.36\% \\
    Informativeness Search & 46.57\% & 33.57\% \\
    Stepwise Beam Search (w/ SG) & 50.49\% & 32.86\% \\
    Informativeness Search (w/ SG) & 51.96\% & 33.93\% \\
    \midrule
    \rowcolor[HTML]{E1E1E1} \multicolumn{3}{c}{\textit{Temperature Sampling}} \\
    Stepwise Beam Search & 41.67\% & 29.64\% \\
    Informativeness Search & 44.12\% & 31.43\% \\
    Stepwise Beam Search (w/ SG) & 47.55\% & 29.64\% \\
    Informativeness Search (w/ SG) & 48.53\% & 32.50\% \\
    \bottomrule
    \end{tabular}
    }
    \caption{Different candidate step generation methods.}
    \label{candidate_generation}
\end{table}

\begin{figure}
    \centering
    \includegraphics[width=0.96\linewidth]{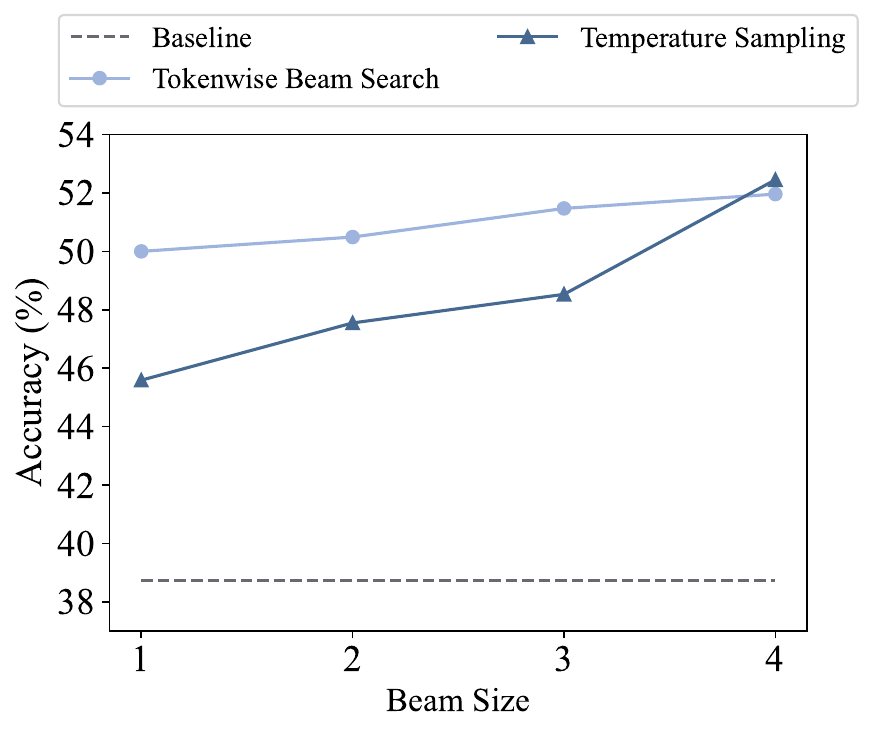}
    \caption{The impact of beam size on our utility-based search for the FOLIO dataset on Llama3.2-3B-Instruct.}
    \label{fig:beam_size}
\end{figure}

We further evaluate the impact of varying beam sizes in our informativeness search, using both tokenwise beam search and temperature sampling for candidate step generation. 
Specifically, we set the sample size to 2 and vary the beam size from 1 to 4. As shown in Fig.~\ref{fig:beam_size}, both alternatives consistently outperform the few-shot CoT baseline. Additionally, our informativeness search continues to improve as beam size increases. Notably, when the search space is constrained (i.e., with a smaller beam size), tokenwise beam search performs better. Based on these findings, we adopt tokenwise beam search for all stepwise beam search methods in our reported results  (Table~\ref{overall_result}$\sim$~\ref{r1_llama_tokens}) considering its better performance and accelerated computational speed.

\subsection{Comparison to Tokenwise Beam Search}
We further compare our informativeness search (beam size $N=3$, sample size $k=2$) with naive tokenwise beam search for whole rationale generation using beam size 3 and 6. Table~\ref{comparison_bs} demonstrate the effectiveness of our method.
\begin{table}[!th]
    \centering
    \setlength\tabcolsep{2pt}
    \resizebox{0.5\textwidth}{!}{
    \begin{tabular}{c|cccc}
    \toprule
    Method &  FOLIO & ProofWriter & MMLU-Pro & GPQA-D \\
    \midrule
    Few-shot CoT & 38.73\% & 40.00\% & 28.57\% & 21.72\% \\
    Tokenwise BS (3) & 43.63\% & 45.00\% & 28.93\% & 21.72\% \\
    Tokenwise BS (6) & 46.08\% & 42.17\% & 31.07\% & 19.19\%\\
    \rowcolor[HTML]{E1E1E1} Informativeness Search & 46.57\% & 50.33\% & 33.93\% & 27.27\% \\
    \bottomrule
    \end{tabular}
    }
    \caption{Comparison with tokenwise beam search using Llama3.2-3B-Instruct for whole rationale generation. Numbers in parentheses denote the beam size.}
    \label{comparison_bs}
\end{table}

\section{Framework Prompts}
\label{prompts}
Table~\ref{prompt_no_grounding_folio},~\ref{prompt_no_grounding_proofwriter},~\ref{prompt_no_grounding_mmlu} and~\ref{prompt_no_grounding_gpqa} present the prompts used in our informativeness search framework without self-grounding strategy for the FOLIO, ProofWriter, MMLU-Pro and GPQA-Diamond datasets. For illustration, Table~\ref{prompt_grounding_gpqa} provides the prompt used in our informativeness search framework with self-grounding strategy on GPQA-Diamond~\footnote{We use GPT-4o and Claude to adjust prompts manually.}.

\section{Illustration of the Grounding Challenge}
\begin{figure}[ht]
    \centering
    \includegraphics[width=0.98\linewidth]{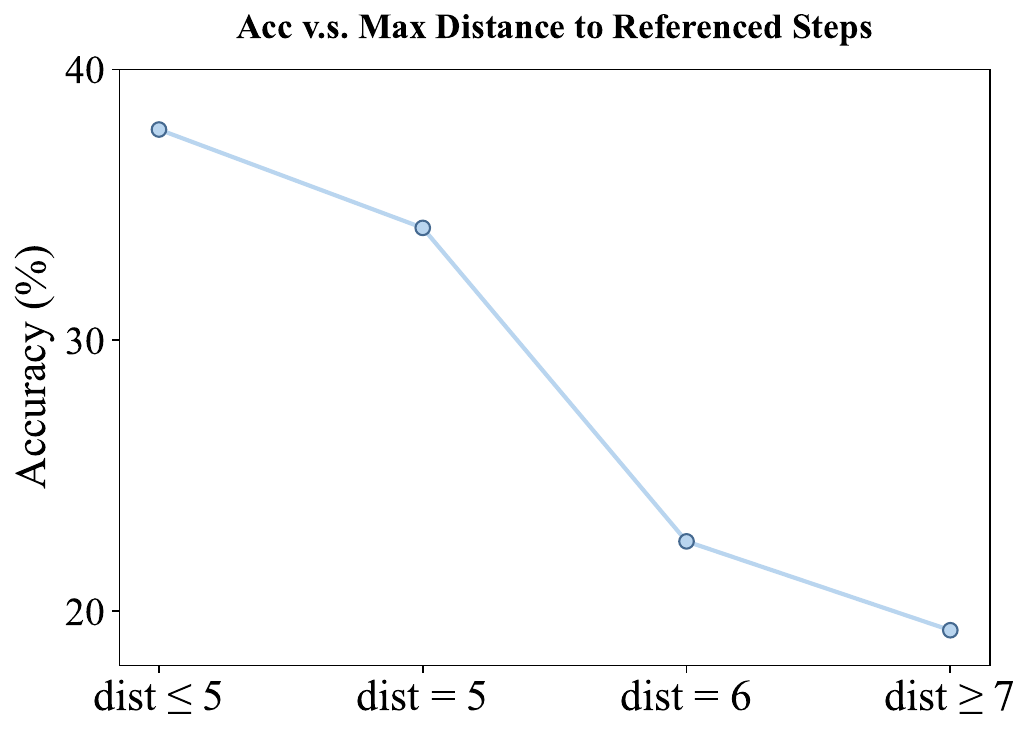}
    \caption{Accuracy versus maximum distance between referencing and referenced steps on CLUTRR.}
    \label{fig:acc_distance}
\end{figure}

We provide a detailed illustration of the challenge LLM face in  grounding prior reasoning steps. Specifically, we analyze all instances involving 8-9 reasoning steps from CLUTRR~\cite{sinha2019clutrr}, a dataset with well-annotated rationales. We evaluate the performance of Llama3-8B-Instruct
across instances with varying maximum distances between referencing and referenced steps. As shown in Fig.~\ref{fig:acc_distance}, performance degrades as the distance to the referenced prior steps grows. This demonstrate the inherent difficulty of grounding prior step, with longer distances (steps accumulating) making grounding progressively harder.

\begin{table*}
\centering
\resizebox{0.98\textwidth}{!}{
\begin{tcolorbox}[colback=blue!5!white,colframe=black!55!black,width=0.98\textwidth,title={Prompt without Self-Grounding (FOLIO)}]
\small
You are a helpful assistant.\\
You will receive a query. Your task is to answer the query.\\
\\
\#\#\#\# Examples\\
Query: LanguageA is a universal language. If a universal language exists, then for every two people if they both know the same universal language they can communicate. Katya cannot communicate with Danil. Katya knows LanguageA. Based on the above information, is the following statement true, false, or uncertain? Danil knows LanguageA.\\
Thought: \\
Because LanguageA is a universal language, and if a universal language exists, then for every two people if they both know the same universal language they can communicate, so every two people that know LanguageA can communicate.\\
Because every two people that know LanguageA can communicate, and Katya knows LanguageA, so Katya can communicate with others that know LanguageA.\\
Because Katya can communicate with others that knows LanguageA, and Katya cannot communicate with Danil, so Danil does not know LanguageA.\\
Therefore, the statement "Danil knows LanguageA." is False.\\
END.\\\
So the answer is: False.\\
------\\
Query: All eels are fish. No fish are plants. A thing is either a plant or animal. Nothing that breathes is paper. All animals breathe. If a sea eel is either an eel or a plant, then a sea eel is an eel or an animal. Based on the above information, is the following statement true, false, or uncertain? Sea eel breathes or is paper.\\
Thought: \\
Because all eels are fish, so a sea eel is a fish.\\
Because no fish are plants, a thing is either a plant or animal, so a fish is an animal. \\
Because a sea eel is a fish, and a fish is an animal, so a sea eel is an animal.\\
Because a sea eel is an animal, and all animals breathe, so a sea eel breathes.\\
Because a sea eel breathes and nothing that breathes is paper, so a sea eel is not paper.\\
Therefore, the statement "Sea eel breathes or is paper." is True.\\
END.\\
So the answer is: True.\\
\\
\#\#\#\# Here's what you need to do. Please first think step-by-step, give out each of your step in a newline, then end your thought with "END.". Finally respond "True", "False" or "Uncertain" in a newline, strictly starting with "So the answer is: ".
\end{tcolorbox}
}
\caption{The prompt without self-grounding on FOLIO.}
\label{prompt_no_grounding_folio}
\end{table*}

\begin{table*}
\centering
\resizebox{0.98\textwidth}{!}{
\begin{tcolorbox}[colback=blue!5!white,colframe=black!55!black,width=0.98\textwidth,title={Prompt without Self-Grounding (ProofWriter)}]
\small
You are a helpful assitant.\\
You will receive a query. Your task is to answer the query.\\
\\
\#\#\#\# Examples\\
Query: Bob is big. Dave is big. Dave is rough. Erin is nice. Erin is white. Gary is nice. Gary is white. Red things are white. All big things are green. All red, white things are nice. All green things are blue. If something is nice then it is big. All blue, green things are rough. All rough things are red. If something is blue then it is nice. If something is red then it is blue. Based on the above information, is the following statement true, false, or unknown? Gary is not red.\\
Thought: \\
Because Gary is nice, and if something is nice then it is big, so Gary is big.\\
Because Gary is big and all big things are green, so Gary is green.\\
Because Gary is green and all green things are blue, so Gary is blue.\\
Because Gary is green and Gary is blue, and all blue, green things are rough, so Gary is rough.\\
Because Gary is rough and all rough things are red, so Gary is red. \\
Therefore, the statement "Gary is not red." is false.\\
END.\\
So the answer is: False.\\
------\\
Query: Anne is nice. Anne is smart. Charlie is green. Fiona is nice. Fiona is round. Fiona is white. Harry is blue. White, kind things are nice. If something is smart and kind then it is green. If something is round and kind then it is white. Smart things are kind. Nice, white things are kind. Round things are kind. If something is nice then it is smart. All white things are round. If Charlie is green then Charlie is white. Based on the above information, is the following statement true, false, or unknown? Charlie is smart.\\
Thought: \\
Because Charlie is green, and if Charlie is green then Charlie is white, so Charlie is white.\\
Because Charlie is white and all white things are round, so Charlie is round. \\
Because Charlie is round and round things are kind, so Charlie is kind. \\
Because Charlie is white and Charlie is kind, and white, kind things are nice, so Charlie is nice.\\
Because Charlie is nice, and if something is nice then it is smart, so Charlie is smart. \\
Therefore, the statement "Charlie is smart." is true.\\
END.\\
So the answer is: True.\\
\\
\#\#\#\# Here's what you need to do. Please first think step-by-step, give out each of your step in a newline, then end your thought with "END.". Finally respond "True", "False" or "Unknown" in a newline, strictly starting with "So the answer is: ".
\end{tcolorbox}
}
\caption{The prompt without self-grounding on ProofWriter.}
\label{prompt_no_grounding_proofwriter}
\end{table*}

\begin{table*}
\centering
\resizebox{0.98\textwidth}{!}{
\begin{tcolorbox}[colback=blue!5!white,colframe=black!55!black,width=0.98\textwidth,title={Prompt without Self-Grounding (MMLU-Pro)}]
\small
You will receive a query and ten options. Your task is to select an option to answer the query.\\
\\
\#\#\#\# Examples\\
Query: Kylar went to the store to buy glasses for his new apartment. One glass costs \$5, but every second glass costs only 60\% of the price. Kylar wants to buy 16 glasses. How much does he need to pay for them?\\
Options: A.24, B.54, C.40, D.32, E.64, F.8, G.16, H.60, I.100, J.74\\
Thought: \\
Because one glass costs \$5, and every second glass costs only 60\% of the price, so the discount price of every second glass is 60/100 * 5 = \$3.\\
Because every second glass is discounted at \$3, and Kylar wants to buy 16 glasses, so Kylar is going to buy 16 / 2 = 8 discounted glasses and 16 - 8 = 8 regular-priced glasses.\\
Because Kylar is going to buy 8 discounted glasses, and every discounted glass is \$3, so Kylar is going to pay 8 * 3 = \$24.\\
Because Kylar is also going to buy 8 regular-priced glasses, and one glass costs \$5, so Kylar will pay 8 * 5 = \$40.\\
Because Kylar will pay \$24 for 8 discounted glasses, and \$40 for 8 regular-priced glasses, so in total Kylar needs to pay 24 + 40 = \$64 for the glasses he wants to buy.\\
END.\\
So the answer is: E.\\
------\\
Query: A refracting telescope consists of two converging lenses separated by 100 cm. The eye-piece lens has a focal length of 20 cm. The angular magnification of the telescope is ?\\
Options: A.10, B.40, C.6, D.25, E.15, F.50, G.30, H.4, I.5, J.20\\
Thought: \\
Because in a refracting telescope both lenses are converging, so their focus must be between the two lenses.\\
Because the focus of both lenses must lie between them, so their focal lengths must add up to their separation. \\
Because the two lenses are separated by 100 cm, and one lens has a focal length of 20 cm, so the other lens must have a focal length of 80 cm.\\
Because one lens has a focal length of 20 cm and the other 80 cm, so the magnification is the ratio of their focal lengths, which is 4.\\
END.\\
So the answer is: H. \\
\\
\#\#\#\# Here's what you need to do. Please first think step-by-step, presenting each of your step in a new line. Then end your thought with "END.". Finally respond with an option from "A", "B", "C", "D", "E", "F", "G", "H", "I" or "J" in a newline, strictly starting with "So the answer is: ".
\end{tcolorbox}
}
\caption{The prompt without self-grounding on MMLU-Pro.}
\label{prompt_no_grounding_mmlu}
\end{table*}

\begin{table*}
\centering
\resizebox{0.98\textwidth}{!}{
\begin{tcolorbox}[colback=blue!5!white,colframe=black!55!black,width=0.98\textwidth,title={Prompt without Self-Grounding (GPQA-Diamond)}]
\small
You will receive a query along with four options. Your task is to select an option to answer the query.\\
\\
\#\#\#\# Examples\\
Query: Kylar went to the store to buy glasses for his new apartment. One glass costs \$5, but every second glass costs only 60\% of the price. Kylar wants to buy 16 glasses. How much does he need to pay for them?\\
Options: \\
(A) 24 \\
(B) 54 \\
(C) 40 \\
(D) 64\\
Thought: \\
Because one glass costs \$5, and every second glass costs only 60\% of the price, so the discount price of every second glass is 60/100 * 5 = \$3.\\
Because every second glass is discounted at \$3, and Kylar wants to buy 16 glasses, so Kylar is going to buy 16 / 2 = 8 discounted glasses and 16 - 8 = 8 regular-priced glasses.\\
Because Kylar is going to buy 8 discounted glasses, and every discounted glass is \$3, so Kylar is going to pay 8 * 3 = \$24.\\
Because Kylar is also going to buy 8 regular-priced glasses, and one glass costs \$5, so Kylar will pay 8 * 5 = \$40.\\
Because Kylar will pay \$24 for 8 discounted glasses, and \$40 for 8 regular-priced glasses, so in total Kylar needs to pay 24 + 40 = \$64 for the glasses he wants to buy.\\
END.\\
So the answer is: D.\\
------\\
Query: A refracting telescope consists of two converging lenses separated by 100 cm. The eye-piece lens has a focal length of 20 cm. The angular magnification of the telescope is ?\\
Options: \\
(A) 10 \\
(B) 6 \\
(C) 4 \\
(D) 25\\
Thought: \\
Because in a refracting telescope both lenses are converging, so their focus must be between the two lenses.\\
Because the focus of both lenses must lie between them, so their focal lengths must add up to their separation. \\
Because the two lenses are separated by 100 cm, and one lens has a focal length of 20 cm, so the other lens must have a focal length of 80 cm.\\
Because one lens has a focal length of 20 cm and the other 80 cm, so the magnification is the ratio of their focal lengths, which is 4.\\
END.\\
So the answer is: C. \\
\\
\#\#\#\# Here's what you need to do. Please first think step-by-step, give out each of your step in a newline. Then end all your thought with "END.". Finally respond with an option from "A", "B", "C" or "D" in a newline, strictly starting with "So the answer is: ".
\end{tcolorbox}
}
\caption{The prompt without self-grounding on GPQA-Diamond.}
\label{prompt_no_grounding_gpqa}
\end{table*}

\begin{table*}
\centering
\resizebox{0.98\textwidth}{!}{
\begin{tcolorbox}[colback=blue!5!white,colframe=black!55!black,width=0.98\textwidth,title={Prompt with Self-Grounding (GPQA-Diamond)}]
\small
You will receive a query along with four options. Your task is to select an option to answer the query.\\
\\
\#\#\#\# Examples\\
Query: Kylar went to the store to buy glasses for his new apartment. One glass costs \$5, but every second glass costs only 60\% of the price. Kylar wants to buy 16 glasses. How much does he need to pay for them?\\
Options: \\
(A) 24 \\
(B) 54 \\
(C) 40 \\
(D) 64\\
Thought: \\
\text{[Step-1]} From Query, because one glass costs \$5, and every second glass costs only 60\% of the price, so the discount price of every second glass is 60/100 * 5 = \$3. \\
\text{[Step-2]} From Step-1 and Query, because every second glass is discounted at \$3, and Kylar wants to buy 16 glasses, so Kylar is going to buy 16 / 2 = 8 discounted glasses and 16 - 8 = 8 regular-priced glasses. \\
\text{[Step-3]} From Step-1 and Step-2, because Kylar is going to buy 8 discounted glasses, and every discounted glass is \$3, so Kylar is going to pay 8 * 3 = \$24. \\
\text{[Step-4]} From Step-2 and Query, because Kylar is also going to buy 8 regular-priced glasses, and one glass costs \$5, so Kylar will pay 8 * 5 = \$40. \\
\text{[Step-5]} From Step-3 and Step-4, because Kylar will pay \$24 for 8 discounted glasses, and \$40 for 8 regular-priced glasses, so in total Kylar needs to pay 24 + 40 = \$64 for the glasses he wants to buy. \\
END.\\
So the answer is: D. \\
------\\
Query: A refracting telescope consists of two converging lenses separated by 100 cm. The eye-piece lens has a focal length of 20 cm. The angular magnification of the telescope is ?\\
Options: \\
(A) 10 \\
(B) 6 \\
(C) 4 \\
(D) 25\\
Thought: \\
\text{[Step-1]} From Query, because in a refracting telescope both lenses are converging, so their focus must be between the two lenses. \\
\text{[Step-2]} From Step-1, because the focus of both lenses must lie between them, so their focal lengths must add up to their separation. \\
\text{[Step-3]} From Step-2 and Query, because the two lenses are separated by 100 cm, and one lens has a focal length of 20 cm, so the other lens must have a focal length of 80 cm. \\
\text{[Step-4]} From Step-3 and Query, because one lens has a focal length of 20 cm and the other 80 cm, so the magnification is the ratio of their focal lengths, which is 4. \\
END.\\
So the answer is: C. \\
\\
\#\#\#\# Here's what you need to do. Please first think step-by-step, give out each of your step in a newline starting with \text{[Step-i]}, and cite the sources (e.g., Step-i, Query) of your premises at the beginning of each step. Then end all your thought with "END.". Finally respond with an option from "A", "B", "C" or "D" in a newline, strictly starting with "So the answer is: ".
\end{tcolorbox}
}
\caption{The prompt with self-grounding on GPQA-Diamond.}
\label{prompt_grounding_gpqa}
\end{table*}

\end{document}